\documentclass[runningheads]{llncs}
\usepackage{graphicx}

\usepackage{tikz}
\usepackage{comment}
\usepackage{amsmath,amssymb} 
\usepackage{color}
\usepackage{caption}
\usepackage{subcaption}

\usepackage[flushleft]{threeparttable}
\usepackage[accsupp]{axessibility}  

\begin{document}
\pagestyle{headings}
\mainmatter
\def\ECCVSubNumber{117}  

\title{DEArt: Dataset of European Art} 


\titlerunning{DEArt: Dataset of European Art}
%
\author{Artem Reshetnikov\inst{1}\orcidID{0000-0003-3257-8512} \and
Maria-Cristina Marinescu\inst{1}\orcidID{0000-0002-6978-2974} \and
Joaquim More Lopez\inst{1}\orcidID{0000-0001-5432-0657}}
\authorrunning{A.Reshetnikov et al.}
%
\institute{Barcelona Supercomputing Center, Barcelona, Spain
\email{\{artem.reshetnikov, joaquim.morelopez, maria.marinescu\}@bsc.es}\\
 }
\maketitle

\begin{abstract}
Large datasets that were made publicly available to the research community over the last 20 years have been a key enabling factor for the advances in deep learning algorithms for NLP or computer vision. These datasets are generally pairs of aligned image / manually annotated metadata, where images are photographs of everyday life. Scholarly and historical content, on the other hand, treat subjects that are not necessarily popular to a general audience, they may not always contain a large number of data points, and new data may be difficult or impossible to collect. Some exceptions do exist, for instance, scientific or health data, but this is not the case for cultural heritage (CH). The poor performance of the best models in computer vision - when tested over artworks - coupled with the lack of extensively annotated datasets for CH, and the fact that artwork images depict objects and actions not captured by photographs, indicate that a CH-specific dataset would be highly valuable for this community. We propose DEArt, at this point primarily an object detection and pose classification dataset meant to be a reference for paintings between the XIIth and the XVIIIth centuries. It contains more than 15000 images, about 80\% non-iconic, aligned with manual annotations for the bounding boxes identifying all instances of 69 classes as well as 12 possible poses for boxes identifying human-like objects. Of these, more than 50 classes are CH-specific and thus do not appear in other datasets; these reflect imaginary beings, symbolic entities and other categories related to art. Additionally, existing datasets do not include pose annotations.  Our results show that object detectors for the cultural heritage domain can achieve a level of precision comparable to state-of-art models for generic images via transfer learning.

\end{abstract}

\section{Introduction}
\label{intro}

Cultural heritage (CH) is important not only for historians or cultural and art institutions. This is an area that permeates society and lends itself to research education and cultural or social projects. Tapping deeper into the richness and potential of CH rests on making it explainable and accessible, and requires efficient indexing and search capabilities. These are built on the premise that quality metadata exists and is available. GLAM (Gallery, Library, Archive, Museum) institutions have been annotating CH artefacts with rich metadata for a long time, but their approach has two shortcomings: (1) the annotations are generated manually as part of a slow and laborious process, and (2) the annotations are usually about the context and making of the artifact or its author.  The assumption is that one sees the object and there is little need to describe it visually. This isn’t always true for people (e.g. in the case of the visually impaired) and it is certainly false when information is needed for automatic consumption by computers. Enabling visual content annotations via an automatic recommendation process could address both these problems.  A good way to approach this challenge is to develop deep learning models for image classification, object detection, or scene understanding. The precision of such models rests on the existence of a large quality dataset, annotated for the target task. Datasets such as MS COCO\cite{mscoco} or Pascal VOC\cite{voc} exist and many deep learning algorithms have been developed that return very good results for photographs of everyday life. When testing these models on artworks, the precision of object detection drops significantly (Section~\ref{experiments}). This is due to reasons of both form and content: paintings are executed in many styles, they have been created to transmit meaning, they capture symbols, depict imaginary beings or artifacts, and reflect objects not in use anymore or actions that we don’t often see in photographs. Additionally, there exists an inherent limitation of the number of existing CH artifacts as opposed to the conceptually unbounded number of photographs that can be taken. All these factors make the existing approaches inadequate for CH. This paper introduces the largest and most comprehensive object detection dataset for CH with annotated poses. To the extent of our knowledge, the most extensive CH  dataset to date has about 6000 images and annotates 7 classes\cite{Garca2018HowTR}\cite{Gonthier2018WeaklySO}\cite{Westlake2016DetectingPI}. Our dataset has 15K images of European paintings between the XIIth and the XVIIIth century. We manually annotate all objects (labels and bounding boxes) corresponding to instances of our 69 classes. Our object detection model has a precision of 31.2\%, which is within 87\% of the state-of-the-art models for photographs. To reach this precision we use transfer learning based on the  Faster RCNN\cite{Ren2015FasterRT} model trained on the MS COCO dataset\cite{mscoco}. In addition to class labels, bounding boxes representing human-like objects also include pose labels, for all the images in our dataset. This is a type of information that identifies an important subset of human-related action verbs. Our paper makes the following main contributions: (1) a dataset of 15K images of paintings between the XIIth-XVIIIth century, annotated for object detection and pose classification; (2) a proof-of-concept that an object detection model can achieve an average precision that is close to that of state-of-the-art models for other domains, and (3) an extensive evaluation, including comparisons with SotA models in computer vision and existing object detections models for cultural heritage. We will make the dataset and model available in August 2022 (DOI:10.5281/zenodo.6984525).      

\section{Related work}

The publishing of large datasets such as Pascal VOC\cite{voc}, MS COCO\cite{mscoco}, or  Open Images\cite{Kuznetsova2020TheOI} enabled researchers in computer vision to improve the performance of the detection of basic object categories in real-life photographs. MS  COCO\cite{mscoco} contains images of objects in their natural environments, which makes this dataset amenable for application in different domains such as video surveillance, security, mobile applications, face and people recognition, etc; neither of these datasets performs well when applied to CH images.  The problem of object detection for paintings has not been studied extensively and is inherently more difficult due to issues such as a big variation in style, the lack of annotated data, and a small dataset by deep learning standards. Artworks often contain objects that are not present in (recent) everyday life; additionally, CH needs more precise annotations than just broad classes. For instance, we need to know that a person is a monk, a king, or a knight to even start understanding the content or meaning of a scene; a person would not serve the purpose. In reality, what we primarily find in CH is research that is based on restricting the object detection problem to a single class, namely a person.  Westlake et al. (2016)\cite{Westlake2016DetectingPI} proposed to perform people detection in a wide variety of artworks (through the newly introduced PeopleArt dataset) by fine-tuning a network in a supervised way. PeopleArt includes 4821 annotated images.  Ginosar et al. (2014)\cite{Ginosar2014DetectingPI} used CNNs for the detection of people in cubist artworks; the authors used 218 paintings of Picasso (a subset of PeopleArt) to fine-tune a pre-trained model. Yarlagadda et al. (2010)\cite{Yarlagadda2010RecognitionAA} collected a dataset for the detection of different types of crowns in medieval manuscripts. More recent approaches for object detection in CH are based on weakly supervised learning or use other ways to generate synthetic data. Gonthier et al. (2018)\cite{Gonthier2018WeaklySO} introduce IconArt, a  new dataset with 5955 paintings extracted from WikiCommons (of which 4458  are annotated for object detection) that annotate 7 classes of objects; their approach is based on a (weakly supervised) multiple-instances learning method.  Kadish et al. (2021)\cite{Kadish2021ImprovingOD} propose to use style transfer (using AdaIn\cite{Huang2017ArbitraryST}) to generate a synthetic dataset that recognizes the class person starting with 61360  images from the COCO dataset. Both IconArt\cite{Gonthier2018WeaklySO} and PeopleArt\cite{Westlake2016DetectingPI} have important weaknesses: they don’t represent significant variations in the representation of class instances, and the list of classes is not representative of the objects represented in cultural heritage. Lastly, Marco Fiorucci et al. (2020)’s\cite{Fiorucci2020MachineLF} survey identifies PrintArt, introduced by Carniero et al. (2012)\cite{Carneiro2012ArtisticIC}, which contains 988  images and identifies 8 classes.  Several other computer vision datasets for cultural heritage exist, such as  SemArt (Garcia et al.(2018))\cite{Garca2018HowTR}, OmniArt (Strezoski et al.(2017))\cite{Strezoski2018OmniArtAL}, ArtDL  (Milani et al. (2020))\cite{Milani2021ADA}, or The Met Dataset (Garcia et al.(2021))\cite{met}. However,  all these datasets focus on image classification or caption generation and they don’t have bounding box annotations that are necessary for object detection. For the task of pose classification from 2D images, most leading approaches are using part detectors to learn action-specific classifiers. Maji et al.\cite{maji2011action} train action-specific pose lets and for each instance create a pose let activation vector that is being classified using SVMs. Hoai et al.\cite{voc} use body-part detectors to localize the human and align feature descriptors and achieve state-of-the-art results on the PASCAL VOC 2012 action classification challenge. Khosla et al.\cite{khosla2014integrating} identify the image regions that distinguish different classes by sampling regions from a dense sampling space and using a random forest algorithm with discriminative classifiers. However, we didn’t find any work related to CH. Our approach is based on using CNN for pose classification.

\section{Dataset}

\subsection{Object categories}
\label{Object_categories}
Selecting the set of CH classes for object detection is a non-trivial exercise; they must be representative for the time period and they should represent a wide range of the possible iconographic meanings. We are exclusively interested in visual classes; this excludes concepts such as thought, intention, mother, etc, given that they are based on assumptions or knowledge not directly apparent from an image. Starting from the MS COCO Dataset\cite{mscoco}, we perform a first step of chronological filtering to keep only the classes whose first known use (per the Merriam Webster dictionary\cite{webster}) is found before the XIXth century; Table~\ref{tab:cocolist} shows these results. Some of them may look incorrect but they are polysemic words with different meanings in the past, e.g. (1) TV: "A sort of annotation in manuscripts" (First known use: 1526) or (2) Keyboard: "A bank of keys on a musical instrument" (First known use: 1776).

\begin{table}
  \caption{MS COCO Dataset classes after chronological filtering}
  \label{tab:cocolist}
  \centering
  \begin{tabular}{|p{\linewidth}|}
    \hline
    'person','car', 'bus', 'train', 'truck',  
    'boat', 'bench',     'bird', 'cat', 'dog', 'horse', 
    'sheep', 'cow', 'elephant', 'bear','zebra','giraffe',
    'umbrella', 'tie', 'skis', 'kite', 'tennis racket', 'bottle', 
    'wine glass', 'cup', 'fork', 'knife', 'spoon', 'bowl', 'banana', 'apple', 
    'sandwich', 'orange', 'broccoli', 'carrot', 'hot dog', 'cake', 'chair', 
    'couch', 'bed', 'dining table', 'toilet', 'tv', 'mouse', 'keyboard', 'oven', 
    'sink', 'refrigerator', 'book', 'clock', 'vase', 'scissors', 'toothbrush' \\
  \hline
  \end{tabular}
  
\end{table}

In a second step, objects not depicted in the present or not real - but present in paintings - must be added to complete the class list. We collect artistic topics which correspond to categories represented in paintings by using Wikimedia Commons as the point of entry. This includes categories that match the regular expression "Paintings of *". For each class C in the filtered MS COCO classes (Table~\ref{tab:cocolist}), we query the Wikimedia API for the category "Paintings\_of\_C" to find out whether it qualifies as a painting class. Table~\ref{tab:cocolistfiltered} shows the MS COCO classes that qualify as painting classes. Most of these classes represent generic individuals and objects devoid of any symbolic or iconographic meaning. To avoid missing a whole range of possible objects which are significant in paintings, we need to include a set of important additional classes. Consider a painting of the crucifixion; Jesus Christ should not be referred to as just a person if we don't want to miss his historical significance and his role in what forms the basis of western iconography. 

\begin{table}
  \caption{MS COCO Dataset classes representing painting classes}
  \label{tab:cocolistfiltered}
  \centering
  \begin{tabular}{|p{\linewidth}|}
    \hline
    'people’, 'trains', 'boats', 'birds', 'cats', 'dogs', 'horses', 'sheep', 'cows', 'elephants', 'bears', 'zebras', 'knives', 'bananas', 'apples', 'oranges', 'chairs', 'beds', 'mice', 'books', 'clocks', 'vases' \\
  \hline
  \end{tabular}
\end{table}

The Wikimedia Commons categories and subcategories are very useful to identify new and richer painting classes by querying for Paintings of\_COCO class\_or its hyponyms; "Paintings of humans" or "Paintings of women" as stand-ins for the MS COCO class person returns categories such as  "Paintings\_of\_ angels\_with\_humans". As a result, angels is added as a new painting class. The pseudocode below illustrates the algorithm of adding new classes starting from an initial class C (in the class list of Table ~\ref{tab:cocolistfiltered}). To avoid generating a huge list of classes in T, we go through a maximum of three nested levels of recursion; this already results in a significant number of classes (963). For this reason, we refine the pseudocode to further filter the classes added to T based on scope and the Wikimedia category hierarchy, as explained below; these cases, including exceptions, are captured in the pseudocode as the predicate filter\_out. 
\begin{table}
  \caption{Classes selection pseudocode}
  \label{tab:code}
  \centering
  \begin{tabular}{|p{\linewidth}|}
    \hline
\noindent
\hspace{0.2cm}$T = \{C\}$

\noindent
\hspace{0.2cm}For each painting class $C \in T$

\noindent
\hspace{0.4cm}Obtain set S of WCo subcategories containing $C$ (sg/pl)

\noindent
\hspace{0.6cm}For each $S_c$ in $S$

\noindent
\hspace{0.8cm}$S = S$/$\{S_c\}$

\noindent
\hspace{0.8cm}If $S_c$ is represented by a string with no ‘\_’ character // no further relationships exist

\noindent
\hspace{1cm}If $S_c \notin T$  and not {\it filter\_out}($S_c$) //  $S_c$ not already inferred from WCo categories

\noindent
\hspace{1.2cm}$ T = T  \cup {S_c}$

\noindent
\hspace{1cm}Otherwise

\noindent
\hspace{1.2cm}$L_S$ = list of all words preceded by ‘\_’ in string representing $S_c$

\noindent
\hspace{1.2cm}For each $K_S \in L_S$

\noindent
\hspace{1.4cm}If $K_S \notin T$  and not {\it filter\_out}($K_S$) // $K_S$  not already inferred from WCo categories

\noindent
\hspace{1.6cm}  $T = T  \cup {K_c}$ \\
  \hline
  \end{tabular}
  
\end{table}
The scope extends to those classes that are found in vocabularies that are commonly used by cultural heritage institutions. Classes were maintained when it was possible to find them in the vocabularies used by one of the following collections: Wikidata, Brill Iconclass AI Test Set, Netherlands Institute for Art History, and the Europeana Entity Collection. Class names that include geographical or positional qualifiers are maintained in form of the unqualified class, e.g. the class kings of Portugal is included as kings (if it isn't already part of the set); likewise, sitting person is retained as person. Categories with less than 10 child subcategories in Wikimedia taxonomy are filtered out\footnote{Other filters may be implemented, e.g. testing the number of artworks in a category rather than the number of subcategories.}. This number is the result of experiments to reach a tradeoff between the final number of classes and covering a sufficiently representative set. Exceptions apply to maintain (a) all categories that are subclasses of MS COCO in Table~\ref{tab:cocolistfiltered} and (b) categories that have less than 10 subcategories but were identified as being of specific interest. The categories of specific interest are those which iconographically characterize a category already included as a class, according to the criteria explained in Section~\ref{Object_categories} , or those that are related to a class that is already included because they are linked or redirected in a linked data database such as Dbpedia. 

A last case is also the most interesting: we do not include classes that refer to complex concepts that may be important in CH due to their symbolic value, for instance recognizable events such as Annunciation, Adoration of the Magi, the Passion of Jesus Christ, etc. The reason for this choice is that there usually exists a big variation in the way that they are represented. We believe that a better way to solve this challenge is at a semantic level, by reasoning about sets of simple objects and their relationships to infer higher-level knowledge. 

As a result of this process we obtain 69 classes that  are either subclasses of a MS COCO superclass (e.g: fisherman - Person; Judith - Person; swan - animal), they are identified with iconographic features, they bear symbolic meaning (e.g: dragon, Pegasus, angel).

\subsection{Pose categories}

The DEArt dataset also contains a classification of poses for human-like creatures. We believe pose classification is useful for several reasons. Being able to automatically detect poses and associate them with bonding boxes is an effective way to create rich data structures - e.g. Knowledge graphs - that can be used for search, browse, inference, or better Query and Answer systems. While descriptions are mostly meant to be consumed by end-users, triples of objects connected by relationships are ideal for machine consumption; poses provide such a subset of action verbs. NLP methods to extract triples (i.e. entities and relationships) from descriptions works best when parsing and analyzing phrases that are simple, otherwise it may wrongly identify objects and/or the relationships between them. At the same time, good image descriptions in natural language tend to be complex; see for instance the descriptions in SemiArt\cite{Garca2018HowTR} or The Prado Museum's website. This makes triple extraction challenging and underlines the importance of automatic pose classification for bounding-boxes. SemiArt\cite{Garca2018HowTR} contains descriptions that add up to more than 3000 statements which include pose verbs in our list. Some of these refer to word meanings which are not pose-related, e.g. "fall in love", "move to [a city]", "the scene is moving", "run a business", or "stand out". We set up an experiment and ran SpaCy\cite{spacy2}with 500 of these statements. Of these, 104 have a subject that is a pronoun or an anaphoric expression (e.g. "the other"), in 49 the subject is wrongly detected (e.g. "Sheltered by this construction stands the throne, with the Madonna and Child and four saints.": subject of stands - sheltered), and 26 detect an incomplete subject. Given that SpaCy\cite{spacy2} parses phrase by phrase, rather than taking into consideration the larger context, it isn't able to resolve co-references. Other more advanced methods such as NeuralCoref\cite{clark2016deep} may be used, although in our experience these don't work fully well for complex language. Regardless of this, the problem of associating poses with bounding boxes stands when there exists more than one bounding box with the same label.

In addition to this challenge, at a higher level of abstraction, the pose of a human-like creature may have symbolic meaning, especially in iconographic images. Different poses may thus give the artwork a very different meaning.

To take the first step towards these goals, we generate a list of relevant poses based on a mix of two approaches: one manual, based on the opinion of cultural heritage specialists, and the second automatic, based on unsupervised learning. By analyzing about 200 random images from our dataset, a cultural heritage expert suggested 12 main pose categories which appear relevant to our dataset and can be interesting for the future enrichment of metadata with actions (see Table~\ref{tab:spec}). 
\begin{table}
  \caption{List of classes suggested by specialists}
  \label{tab:spec}
        \begin{tabular}{|p{\linewidth}|}
        \hline
        'walk/move/run/flee', 'bend', 'sit', 'stand', 'fall', 'push/pull', 
        'throw', 'kneel', 'lift/squat', 'beg/pray', 'bow', 'step on', 'drink/eat', 'lay down' \\
        \hline
        \end{tabular}
  
  \centering
  \caption{List of classes extracted by suggested approach}
  \label{tab:approach}
        \begin{tabular}{|p{\linewidth}|}
        \hline
        'lift/squat', 'move', 'watch', 'falls', 'kneel', 'ride', 'stand', 'eat', 'sit', 'pull/push', 'fly', 'sleep', 'bend', 'beg/pray' \\
        \hline
        \end{tabular}
  \caption{Final list of classes}
  \label{tab:classesfinal}
  \centering

        \begin{tabular}{|p{\linewidth}|}
        \hline
         'bend',' fall',' kneel','lay down', 'partial',' pray',' push/pull',' ride',' sit/eat',
         'squats','stand','walk/move/run', 'unrecognizable' \\
        \hline
        \end{tabular}

\end{table}
Independently, we extracted all the verbs present in the descriptions of the 19245 images in the SemiArt\cite{Garca2018HowTR} dataset. All descriptions were tokenized and, using part-of-speech tagging, we created a pool of verbs from which we chose the 2000 most common. This approach is based on verbs clustering and the intuition is that verbs appear in the same cluster if they are related to the similar actions. Iterative filtering allows us to keep clusters which are related to poses of human-like creatures in artworks. For this purpose, we use pre-trained GloVe\cite{pennington2014glove} vectors to represent verbs in an embedding space so that we can maintain their semantic meaning.  In each iteration we define the best value for the number of clusters based on computing a score for a number clusters between 1 to 500 and choosing the number with the highest score. We then delete clusters which are not suitable for our purpose, i.e. are not about poses and represent no visual image. The best (but also more computationally expensive) approach to defining the optimal number of clusters during each iteration is to use the silhouette score\cite{aranganayagi2007clustering}. 
The silhouette coefficient for an instance is computed as \((b-a) / max(a, b)\), where \(a\) is the mean distance to the other instances in the same cluster (i.e., the mean intra-cluster distance) and \(b\) is the mean nearest-cluster distance (i.e., the mean distance to the instances of the next closest cluster, defined as the one that minimizes b, excluding the instance’s own cluster). The silhouette coefficient can vary between –1 and +1. A coefficient close to +1 means that the instance is well inside its own cluster and far from other clusters, 0 means that it is close to a cluster boundary, and a coefficient close to –1 means that the instance may have been assigned to the wrong cluster.  After 42 iterations we were able to define 14 main clusters that are related to poses of human-like creatures (See Table~\ref{tab:approach})

Our final list of categories of poses is the intersection of the two approaches, to which we add the class “ride”, while we include class “eat” into class “sit” based on the fact that someone who eats is also sitting and almost never actually eating but rather having food before them. We also add the class ”partial” and “unrecognizable” for those pictures where the human-like creature is only partially represented (e.g. a man from the torso up) and for those images where action can’t be recognized (See Table~\ref{tab:classesfinal})

\subsection{Image collection process}
\label{image_collection_process}

\begin{figure}
     \centering
     \begin{subfigure}[b]{0.18\linewidth}
         \centering
         \includegraphics[width=\linewidth]{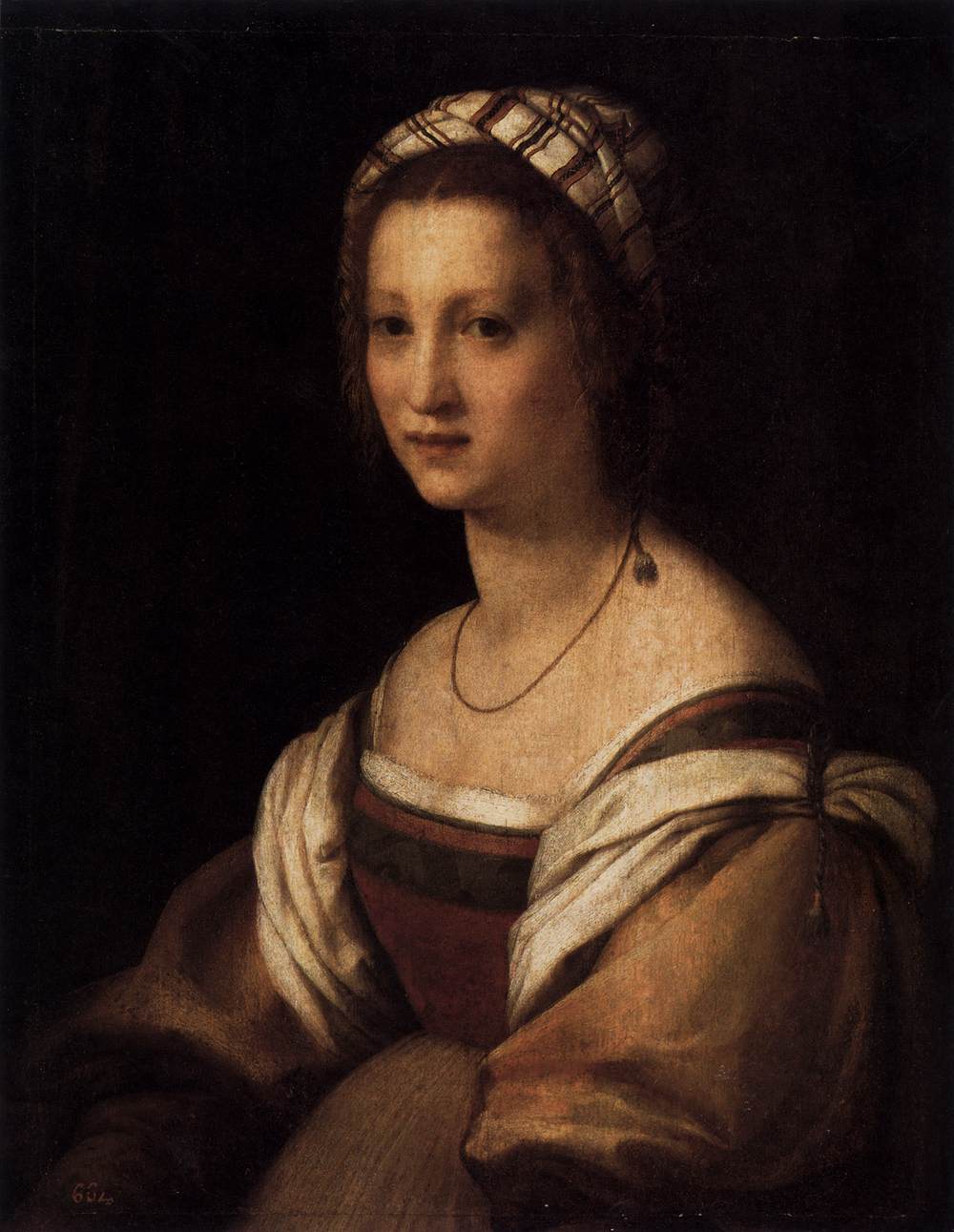}
         \label{fig:y equals x}
     \end{subfigure}
     \hfill
     \begin{subfigure}[b]{0.37\linewidth}
         \centering
         \includegraphics[width=\linewidth]{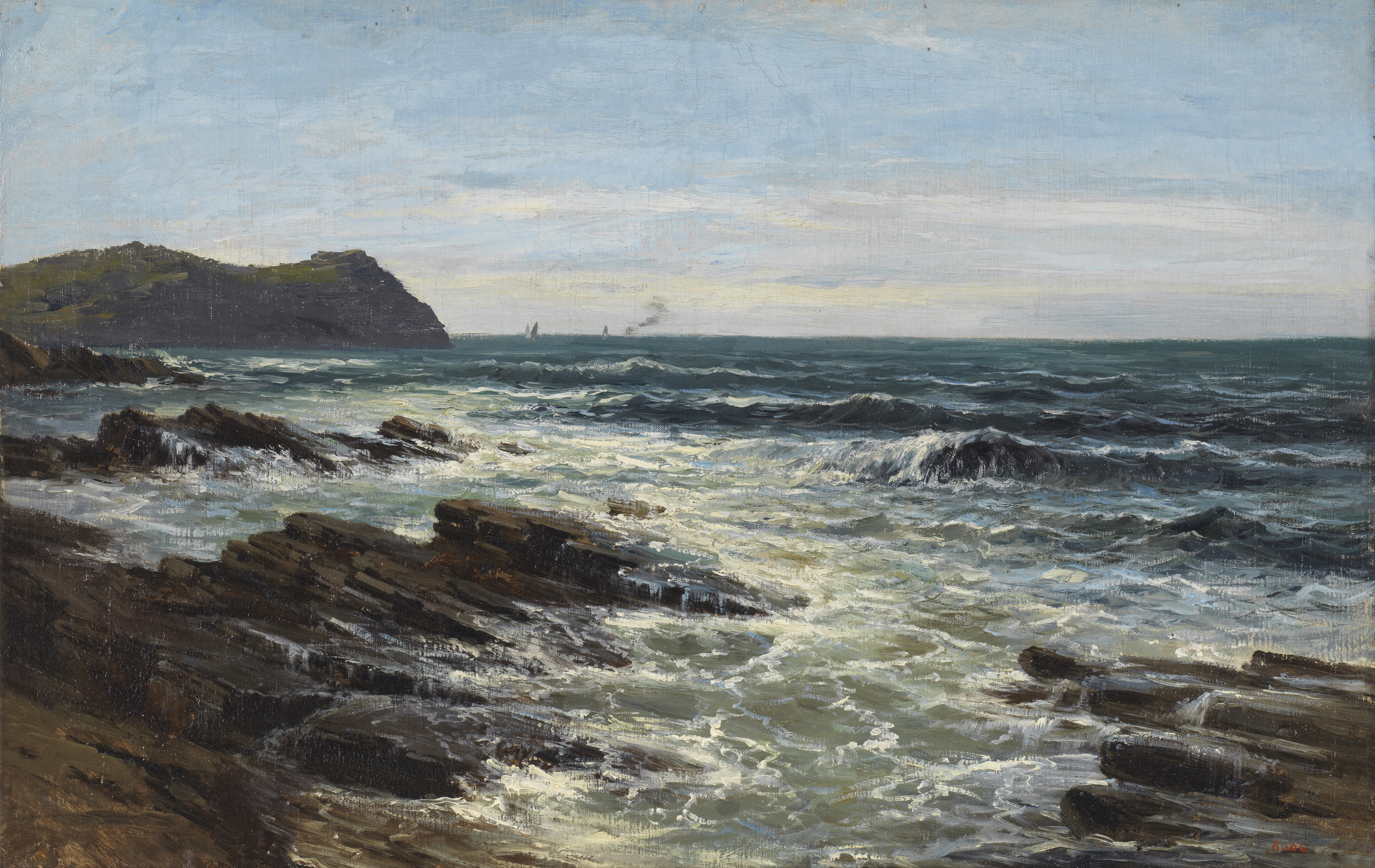}
         \label{fig:three sin x}
     \end{subfigure}
     \hfill
     \begin{subfigure}[b]{0.32\linewidth}
         \centering
         \includegraphics[width=\linewidth]{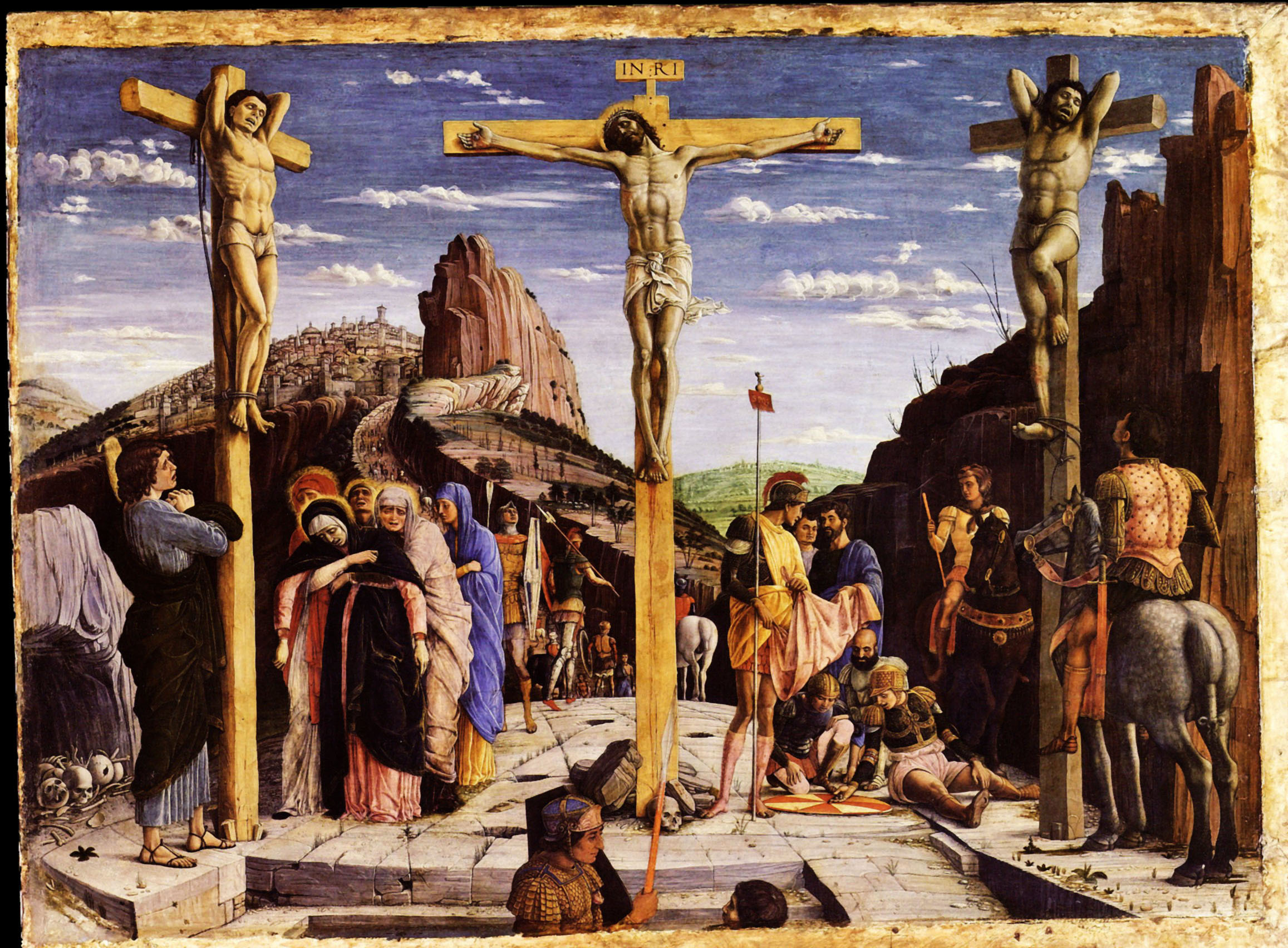}
         \label{fig:five over x}
     \end{subfigure}
        \caption{Examples of Iconic-object, iconic-scene and non-iconic images}
        \label{fig:iconic}
    \label{fig:iconic-noniconic}
\end{figure}

Everyday language defines something to be ”iconic” to mean that it is well-established, widely known, and acknowledged especially for distinctive excellence. Research in computer vision (CV) uses its own concept of ”iconic” to follow the definition introduced by Berg et al. (2009)\cite{Berg2009FindingII}. Typical iconic-object images include a single object in a canonical perspective\cite{canons}. Iconic-scene images are shot from canonical viewpoints\cite{canons} and commonly lack people. Non-iconic images contain several objects with a scene in the background. Figure ~\ref{fig:iconic-noniconic} gives some typical examples. For the rest of this paper, we use the CV terminology when we refer to "iconic". 

\begin{figure}
\centering
\includegraphics[width=\linewidth]{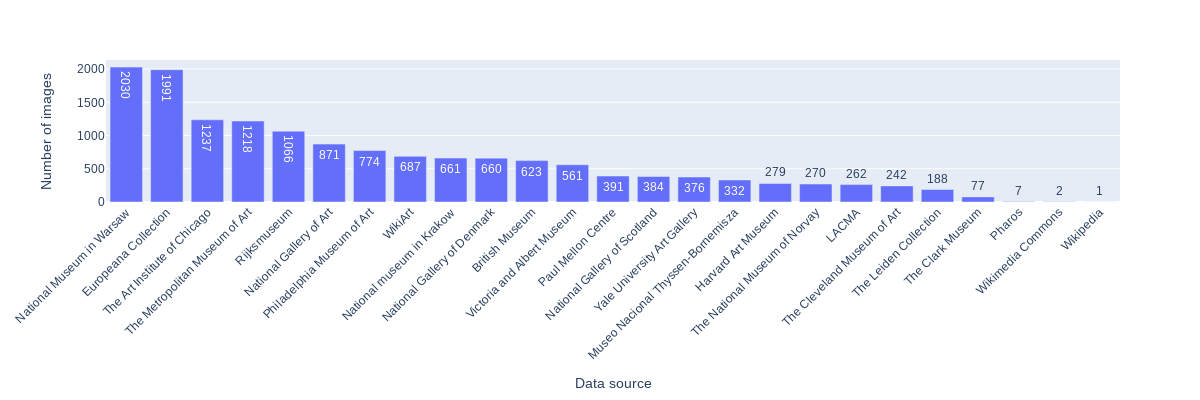}
\caption{Distribution of images by data source}
\label{fig:sources}
\end{figure}

Iconic images are easier to find using search engines or tags on data aggregation platforms, and they are more straightforward to annotate manually. At the same time, they tend to lack the contextual information present in non-iconic images and thus generally represent concepts with no greater complexity or symbolism. In non-iconic images, additional objects play a key role in scene understanding because they bring in an additional layer of symbolic information. To be able to perform inference and iconographic understanding of scenes in the future, our goal is to collect a dataset with a majority of non-iconic images. Our data comes from the following sources: the Europeana Collections, Wikimedia Commons, British Museum, Rijksmuseum, Wikipedia, The Clark Museum, The Cleveland Museum of Art, Harvard Art Museum, Los Angeles County Museum of Art, The Leiden Collection, Paul Mellon Centre, Philadelphia Museum of Art, National Gallery of Scotland, Museo Nacional Thyssen-Bornemisza, Victoria and Albert Museum, National Museum in Warsaw, Yale University Art Gallery, The Metropolitan Museum of Art, National Gallery of Denmark, The National Gallery of Art, The Art Institute of Chicago, WikiArt, PHAROS: The International Consortium of Photo Archives. When possible, in addition to checking the time and place of the creation of the artwork, we filter by license type and only select images open for reuse. Figure~\ref{fig:sources} illustrates the number of images downloaded from each source.

\subsection{Image annotation}

The annotations we are interested in are of two types: bounding boxes with their object labels and pose labels associated with bounding boxes.

We annotate both objects and poses of human-like objects in images following the rules described in "The PASCAL Visual Object Classes (VOC) Challenge"\cite{voc} and set three objectives: (1) Consistency: all annotators should follow the same rules with regard to the list of classes, the position of bounding boxes, and truncated objects. We achieve consistency by exclusively using the project team members (from two institutions) as annotators and following guidelines we drew after extensive discussions. We use LabelMe as the user interface\cite{Russell2007LabelMeAD} for object annotations. (2) Accuracy: the bounding boxes should be tight, and object labels must be from the list of classes. Poses of human-being creatures were classified according to the list of classes. We achieve accuracy by manual checking. (3) Exhaustiveness: all object instances must be labeled. We achieve exhaustiveness by manual checking. 
The object labels and bounding boxes are generated as follows:

The object labels and bounding boxes are generated as follows:
\begin{enumerate}
    \item  An original image selection process chose paintings after the XIIth century until the XVIIIth century. The result is a dataset of 15K images, of which about 80\% are non-iconic. Our research team members annotated 10K images manually; 
 
    \item The rest of 5K was annotated in a semi-supervised manner by using the currently trained model and correcting the annotations manually, in 3 ingestion phases of 2K each, followed by retraining. The training was done over 70\% of the dataset. Semi-supervised learning helps us reduce the time for the annotation process, given that the human annotator does not start from zero, but rather accepts or fixes annotations recommended by the current model. Each data ingestion phase consists of several steps: (1) Collecting data, (2) Detecting objects using the model trained on the previous version of the dataset, (3) Manually correcting bounding boxes to resize them around the objects, eliminate falsely detected objects, rename falsely labeled boxes, and manually annotate undetected objects, and (4) Include corrected data into the dataset and retrain the model for the next ingestion iteration. During the entire annotation process, the dataset is periodically observed to double-check the quality of the annotations. This step is performed approximately every time 2000 new images are added, and it evaluates the 10 classes with the most number of instances: person, tree, boat, angel, halo, nude, bird, horse, book, and cow. The evaluation subset is created by randomly extracting 100 images per category.  

\end{enumerate}

The same process was applied for the annotation of poses of human-like creatures. Using semi-supervised learning, the research team (we) periodically trained a classification model based on annotated data. In later steps, the research team (we)  corrected the generated classes rather than annotating them manually.

\subsection{Dataset statistics}

\begin{figure}
    \begin{subfigure}[b]{\linewidth}
         \centering
         \includegraphics[ width = \linewidth]{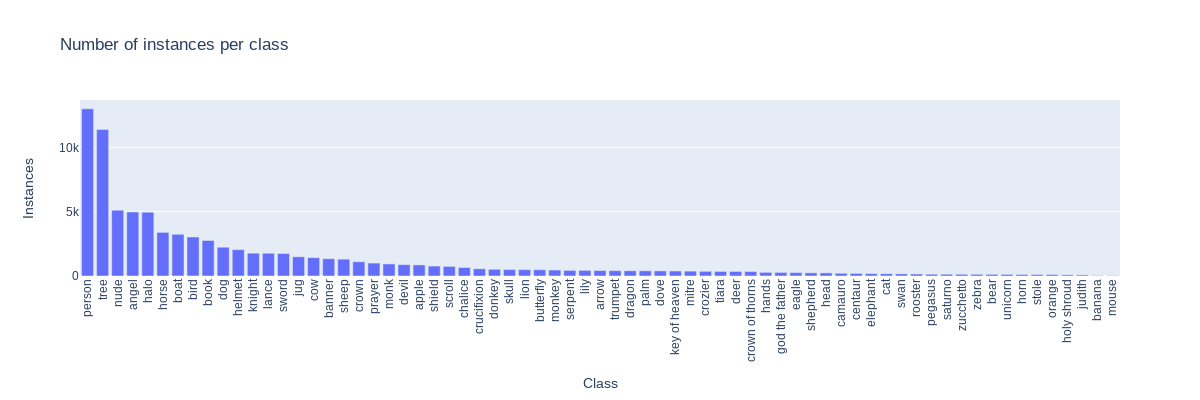}
         \caption{Number of total instances per class}
         \label{fig:classes}
     \end{subfigure}

     \begin{subfigure}[b]{0.45\linewidth}
         \centering
         \includegraphics[width=\linewidth]{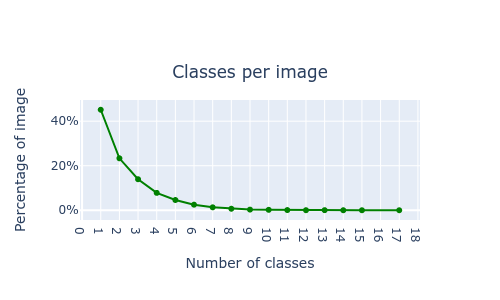}
         \caption{Proportion of the dataset with specific number of classes}
         \label{fig:classes_1}
     \end{subfigure}
     \hfill
     \begin{subfigure}[b]{0.45\linewidth}
         \centering
         \includegraphics[width=\linewidth]{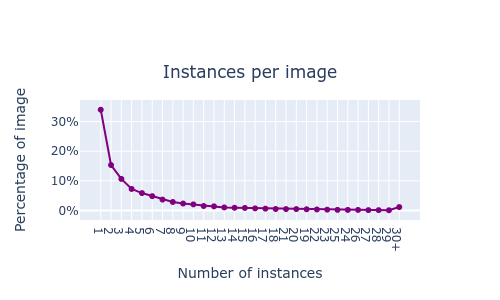}
         \caption{Proportion of the dataset with specific number of instances}
         \label{fig:classes_2}
     \end{subfigure}
        \caption{Dataset statistics}
        \label{fig:stat}
\end{figure}

Figure~\ref{fig:classes} illustrates the total number of instances per class for each of the 69 categories. A class person is by far the most frequent in the dataset, currently with around 46990 instances; the bar for a class person is truncated to make minority classes such as “saturno”, ”stole” or “holy shroud” visible. Given that instances of minority classes are rarely the subject of iconic images but they can appear in non-iconic images, adding complex scene images may help increase the total number of minority instances. 

Figures~\ref{fig:classes_1} and~\ref{fig:classes_2} illustrate the amount of contextual information\cite{canons} in our dataset, which we define as the average number of object classes and instances per image. The figures confirm that during the process of data collection we pay special attention to choosing non-iconic images to be able to obtain a rich dataset. During the annotation process, we made a periodic health check analysis of the dataset to understand which classes were over or under-represented (minority classes). The results directed our search to prioritize images that include minority classes. Differently from photographs, sometimes we have inherent limitations because we cannot produce additional old paintings if they don’t exist. The entire annotation process resulted in 105799 bounding boxes. Additionally, we classified 56230 human-like creatures.

\section{Experiments}
\label{experiments}

\subsection{Object detection}
We implement a deep learning architecture based on Faster R-CNN\cite{Ren2015FasterRT}  to train object detection models over the DEArt dataset. We split our dataset into 70\% training, 15\% validation, and 15\% test sets (Training set: 10500,Validation/Test sets: 2250). The choice of images is random within each class, and we use the annotated-images\footnote{\url{https://github.com/SaberD/annotated-images}} Python library to select images such that these percentages are as closely as possibly met for each of the 69 classes. After comparing several possibilities among the different ImageNet models\cite{imagenet}, we concluded that residual networks (ResNet\cite{He2016DeepRL}) are the best architecture for feature extraction - which coincides with [13]. Taking into account that photograph datasets such as MS COCO\cite{mscoco} contain about 20 times more images, and to increase the final precision of object detection, we decided to use a transfer learning approach. Concretely, we use the Resnet-152 V1\cite{He2016DeepRL} Object detection model, pretrained on the MS COCO 2017\cite{mscoco} dataset. Given that our dataset is in Pascal VOC format, we chose AP@0.5 per class and mAP@0.5\cite{voc} as evaluation metrics.

\begin{figure}

    \centering
    \includegraphics[width=\linewidth]{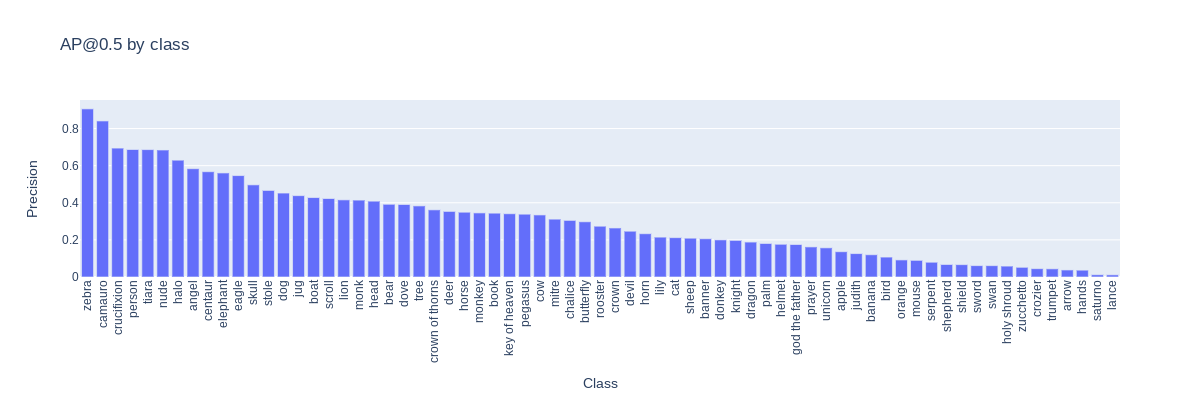}
    \caption{The average precision by class for Faster RCNN trained on DEArt dataset}
    \label{fig:precision_fast}

\end{figure}

Our Faster R-CNN\cite{Ren2015FasterRT} model results in an mAP@0.5 = 31.2; the average precisions by class are presented in Figure~\ref{fig:precision_fast}.

The results of our experiments show that including many complex (non-iconic) images in the training set may not always help when evaluating class by class. Such examples may act as noise and contaminate the model if it isn’t rich enough to capture the variability in a visual representation of the objects, especially if we consider the sizeable number of classes we cover relative to the number of images of the dataset. Furthermore, we compared the results of the evaluation with object detection datasets. Table~\ref{tab:comparing} shows the results of applying three pretrained models (trained on MS COCO, PASCAL VOC, Open Images) over test set of DEArt; the last rows show the results of our the model tained on our dataset. We present AP@0.5 results for the 16 classes that are included both in MS COCO and our dataset. We can clearly see that our model significantly improves the precision of detection of every-day objects as they are depicted in cultural heritage artifacts, when compared to the photograph-based models. At the same time, MS COCO, PASCAL VOC, and Open Images do not cover - and therefore could not recognize - the other 53 cultural heritage-specific classes. On a different note we want to mention some examples of the fall in precision of the CV models when applied to our dataset: for class person, MS COCO goes from 0.36 (when tested over their data) to 0.25 (over our data), while PASCAL VOC goes from 0.22 down to 0.05. For class sheep, MS COCO goes from 0.28 to 0.15 and PASCAL VOC from 0.17 to 0.004. For class horse, 0.44 to 0.2 (MS COCO) and 0.33 to 0.03 (PASCAL VOC).
\begin{table}[bp]
\centering
\caption{Examples for precision of different models (reported in~\cite{mscoco}) when tested over DEArt}
\label{tab:comparing}
\resizebox{12.2cm}{!}{\begin{tabular}{|l|l|l|l|l|l|l|l|l|l|l|l|l|l|l|l|l|}

\hline
            & apple & banana & bear & bird & boat & book    & cat  & cow  & dog  & elephant & horse & mouse & orange & person & sheep & zebra \\ \hline
MS COCO     & 0.04  & 0.008  & 0.03 & 0.12 & 0.24 & 0.05    & 0.04 & 0.23 & 0.12 & 0.21     & 0.2   & 0     & 0.4    & 0.25   & 0.15  & 0.89  \\ \hline
Open Images & 0.008 & 0.005  & 0.12 & 0.01 & 0.07 & 0.00007 & 0.04 & -  & 0.08 & 0.38     & 0.09  & 0     & 0      & 0.07   & 0.04  & 0.5   \\ \hline
Pascal VOC  & -   & -    & -  & 0.02 & 0.05 & -     & 0.09 & 0.13 & 0.02 & -      & 0.03  & -   & -    & 0.05   & 0.004 & -   \\ \hline
DEArt       & 0.13  & 0.12   & 0.39 & 0.15 & 0.42 & 0.34    & 0.21 & 0.33 & 0.45 & 0.56     & 0.34  & 0.09  & 0.09   & 0.68   & 0.2   & 0.91  \\ \hline
\end{tabular}}
\caption{F1 score for pose classification by classes}
\label{tab:poses}
\resizebox{12.2cm}{!}{\begin{tabular}{|l|l|l|l|l|l|l|l|l|l|l|l|l|l|}
\hline
\textbf{Class} & \textit{bend} & \textit{fall} & \textit{kneel} & \textit{lie down} & \textit{partial} & \textit{pray} & \textit{push/pull} & \textit{ride} & \textit{sit/eat} & \textit{squats} & \textit{stand} & \textit{unrecogn.} & \textit{walk} \\ \hline
\textbf{F1}    & 0.33          & 0.08          & 0.32           & 0.80              & 0.90             & 0.33          & 0.09                 & 0.10          & 0.83               & 0.12            & 0.84           & 0.89                     & 0.50          \\ \hline
\end{tabular}}

\end{table}

\subsection{Pose classification}
We built a small version of the Xception\cite{chollet2017xception} network for pose classification. Taking into account that we have a large amount of data, we decided to train the model from scratch using a customized architecture. We used KerasTunner to tune the hyperparameters and architecture, and we split our dataset into 70\% training, 15\% validation, and 15\% test sets. The dataset is highly unbalanced, which made us decide to use the F1 score for the evaluation of the model. Evaluation over all 12 classes shows F1=0.471, with weighted F1=0.89. The minority pose classes are ride, squat, fall, and push/pull(See Table~\ref{tab:poses}). 
\section{Discussion}
\label{Discussion}
\subsection{Poses}
\label{sec:poses}

Rich metadata in machine-readable form is fundamental to enable better or novel functionalities related to search, browse, recommendation, or question answering.  While detection of entities is a relatively common task, learning relationship labels is not. On the other hand, having information about the actions that objects perform or the relationships between them allows the creation of structures such as knowledge graphs, which enable new functionality e.g. inference. A subset of such actions is the set of poses. One may think of these as attributes of the object, while others may consider them as relationships with other objects, e.g. person lies on the bench, angel falls from a cloud, priest prays to an icon, etc. In either case, they qualify objects in new ways that can improve both user experience and machine-exploitable knowledge.

It may be the case that some poses one expects to be detected are not present in our list. This is due to filtering out verbs that are below the  2000 occurrences; while we could play with this number, training a good pose classifier requires a reasonable minimum number of instances. It is conceivable that some of the pose words we eliminated can be in fact very important for the symbolic meaning of artwork despite not being frequent. In this case, we may want to allow the cultural heritage specialist to add pose words to the list and find/produce more aligned image - caption pairs for those minority poses. Lastly, poses further the possibility of drawing symbolic meaning from paintings, especially the iconographic ones.

\subsection{Generating new object labels without annotation }
\label{generating}
While we consider our class set to be representative, it is by no means complete. The list of classes could be extended both in breadth (e.g. including body parts, fish, broom, etc) as well as in depth (e.g. adding refinements of existing classes such as carpenter, beggar, archangel, fisherman, etc). Some of these classes may be very useful in subsequent inference steps to extract symbolic meaning from a painting. Given the relatively small number of CH images, some class set extensions could be useful while others would only make the model less accurate. A way forward to deal with this challenge is to use additional technology to complement deep learning approaches, for instance approaches based on common knowledge. These can allow further refinement of class labels (and thus generate object labels not in the class list without the need of manual annotation) or even the inference of probable relationships between objects in an image.

\subsection{Limitations}

Our dataset targets European cultural heritage imagery, of which we focus in particular on the period between the XIIth and the XVIIIth centuries. We also collected a 10K dataset of European artwork between the XIXth century and now, and we tested our (XII-XVIII) model over this dataset. Section 4.2 shows that it is possible to receive acceptable results. On the other hand, it is hard to know how well or poorly a European art model will generalize when applied to visual arts of other cultures; we expect that not very well. More annotations than only poses would be useful to fully inter-relate the objects detected in a painting, and therefore interconnect multiple bounding boxes. This is part of future work.

\subsubsection*{Acknowledgements} 
This research has been supported by the Saint George on a Bike project 2018-EU-IA-0104, co-financed by the Connecting Europe Facility of the European Union.

%
%
\bibliographystyle{splncs04}
\bibliography{egbib}
\end{document}